\documentclass[10pt,twocolumn,letterpaper]{article}
\usepackage{iccv}
\usepackage{times}
\usepackage{epsfig}
\usepackage{graphicx}
\usepackage{amsmath}
\usepackage{float}
\usepackage{amssymb}
\usepackage{amsfonts}
\usepackage{algorithm}
\usepackage{algorithmic}
\usepackage{subcaption}
\usepackage{booktabs}
\usepackage{lipsum}
\usepackage[symbol*]{footmisc} 
\DefineFNsymbolsTM{myfnsymbols}{
    \textasteriskcentered * 
    \textdagger \dagger 
    \textdaggerdbl \ddagger 
}%
\usepackage{hyperref}
\hypersetup{
    colorlinks=true,
    linkcolor=blue,
    filecolor=magenta,      
    urlcolor=cyan,
}
\setfnsymbol{myfnsymbols} 
\newcommand{\lvl}{~~~~~}

\iccvfinalcopy 
\def\iccvPaperID{1570} 

\ificcvfinal\fi
\begin{document}

\title{Cross-Domain Adaptation for Animal Pose Estimation}
\author{
Jinkun Cao$^{1}$\thanks{Part of this work was done when Jinkun Cao and Hongyang Tang were research interns in Tencent.}~~~
Hongyang Tang$^{1*}$~~~
Hao-Shu Fang$^{1}$~~~
Xiaoyong Shen$^{2}$~~~
Cewu Lu$^{1}$\thanks{Cewu Lu is the corresponding author: lucewu@sjtu.edu.cn}~~\thanks{Cewu Lu is a member of MoE Key Lab of Artificial Intelligence, AI Institute, Shanghai Jiao Tong University}~~~
Yu-Wing Tai$^{2}$
\\
\and
{$^{1}$Shanghai Jiao Tong University~~~~~~~~~~~~~~~~~~~~~~~~$^{2}$Tencent~~~~~~~~~~~~~}
\\
\and
{\tt\small \{caojinkun, lucewu\}@sjtu.edu.cn}\\
\and
{\tt\small \{thutanghy, fhaoshu, goodshenxy\}@gmail.com~~yuwingtai@tencent.com}
}
\maketitle
\begin{abstract}
    In this paper, we are interested in pose estimation of animals. Animals usually exhibit a wide range of variations on poses and there is no available animal pose dataset for training and testing. To address this problem, we build an animal pose dataset to facilitate training and evaluation. Considering the heavy labor needed to label dataset and it is impossible to label data for all concerned animal species, we, therefore, proposed a novel cross-domain adaptation method to transform the animal pose knowledge from labeled animal classes to unlabeled animal classes. We use the modest animal pose dataset to adapt learned knowledge to multiple animals species. Moreover, humans also share skeleton similarities with some animals (especially four-footed mammals). Therefore, the easily available human pose dataset, which is of a much larger scale than our labeled animal dataset, provides important prior knowledge to boost up the performance on animal pose estimation. Experiments show that our proposed method leverages these pieces of prior knowledge well and achieves convincing results on animal pose estimation. The built dataset and other resources have been publicly released on \hyperlink{www.jinkuncao.com/animalpose}{www.jinkuncao.com/animalpose}.
\end{abstract}
    
    \begin{figure*}
        \centering
        \includegraphics[width=17cm]{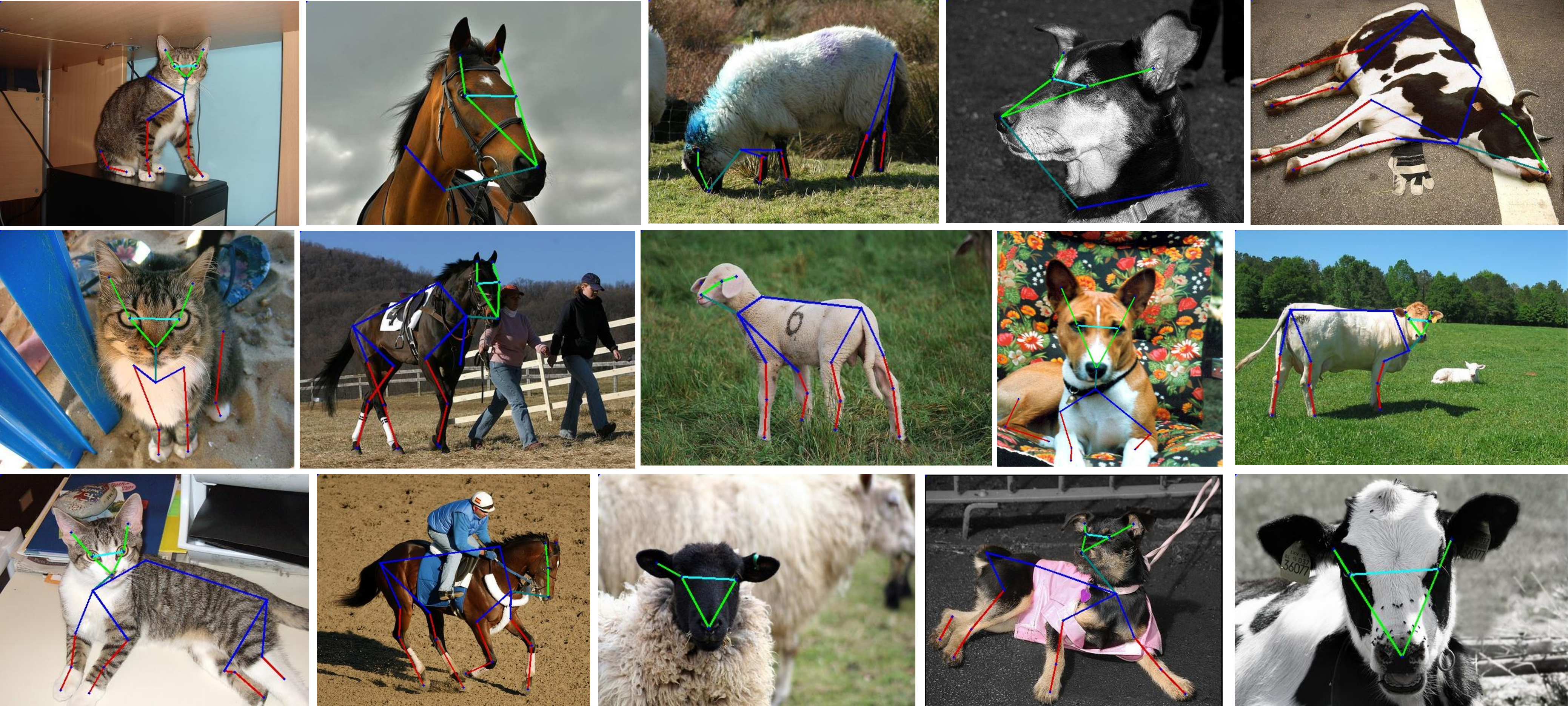}
        \caption{Some samples from the Animal-Pose dataset.}
        \label{animalsample}
    \end{figure*}

\section{Introduction}
    In this paper, we aim to tackle the animal pose estimation problem, which has a wide range of applications in zoology, ecology, biology, and entertainment. Previous works~\cite{alphapose,cao2017realtime, cpn, deeppose} only focused on human pose estimation and achieved promising results. The success of human pose estimation is based on large-scale datasets~\cite{COCO, MPII}. The lack of a well-labeled animal pose dataset makes it extremely difficult for existing methods to achieve competitive performance on animal pose estimation.
    
    In practice, it is impossible to label all types of animals considering there are more than million species of animals and they have different appearances. Thus, we need to exploit some useful prior that can help us to solve this problem, and we have identified three major priors. First, pose similarity between humans and animals or among animals is important supplementary information if we are targeting for four-legged mammals. Second, we already have large-scale datasets (e.g.~\cite{COCO}) of animals with other kinds of annotation which will help to understand animal appearance. Third, considering the anatomical similarities between animals, pose information of a certain class of animals is helpful to estimate animals' pose of other classes if they share a certain degree of similarity.
    
    With the priors above, we propose a novel method to leverage two large-scale datasets, namely pose-labeled human dataset and box-labeled animal dataset, and a small pose-labeled animal dataset to facilitate animal pose estimation. In our method, we begin from a model pretrained on human data, then design a ``weakly- and semi-supervised cross-domain adaptation''(WS-CDA) scheme to better extract cross-domain common features. It consists of three parts: \emph{feature extractor}, \emph{domain discriminator} and the \emph{keypoint estimator}. The \emph{feature extractor} extracts features from input data, based on which the \emph{domain discriminator} tries to distinguish which domain they come from and the \emph{keypoint estimator} predicts keypoints.  With \emph{keypoint estimator} and \emph{domain discriminator} optimized adversarially, the discriminator encourages the network to be adaptive to training data from different domains. This improves pose estimation with cross-domain shared information.
    
    After WS-CDA, the model already has the pose knowledge for some animals. But it still does not perform well on a specific unseen animal class because no supervised knowledge is obtained from this class. Targeting to improve it, we propose a model optimization mechanism called ``Progressive Pseudo-label-based Optimization''(PPLO). The keypoints prediction on animals of new species is optimized using the pseudo-labels which is generated based on selected prediction output by the current model. The insight is that animals of different kinds often share many similarities, such as limb proportion and frequent gesture, providing prior to inferring animal pose. And the prediction with high confidence is expected to be quite close to ground truth, thus bringing augmented data into training with little noise. A self-paced strategy~\cite{selfpacedlearning, selfpacedcurlearning} is adopted to select pseudo-label and to alleviate noise from unreliable pseudo labels. An alternating training approach is designed to encourage model optimization in a progressive way.
    
    We build an animal pose dataset by extending~\cite{poselets} to provide basic knowledge for model training and evaluation.  Five classes of four-legged mammals are included in this dataset:  dog, cat, horse, sheep, cow. To better fuse the pose knowledge from human dataset and animal dataset, the annotation format of pose for this dataset is made easy to be aligned to that of popular human pose dataset\cite{COCO}.
    
    Experimental results show that our approach solves the animal pose estimation problem effectively. Specifically, we achieve 65.7 mAP on test set with a very limited amount of pose-labeled animal data involved in training, close to the state-of-the-art level of accuracy for human pose estimation. And more importantly, our approach gives promising results on cross-domain animal pose estimation, which can achieve 50+ mAP on unseen animal classes without any pose-labeled data for it.

    \section{Related Work}
    
    Pose estimation focuses on predicting body joints on detected objects. Traditional pose estimation is performed on human samples~\cite{COCO, alphapose, 2steppose1, 2steppose2, deeppose}. Some works also focus on the pose of specific body parts, such as hands~\cite{handpose1, handpose2} and face~\cite{headpose1, headpose2, headpose3}. Besides these traditional applications, animal pose estimation brings value in many application scenarios, such as shape modeling~\cite{animalpose}. However, even though some works study the face landmarks of animals~\cite{animalface1, animalface2, animalface3}, the skeleton detection on animals is rarely studied and faces many challenges. And the lack of large-scale annotated animal pose datasets is the first problem to come. Labeling data manually is labor-intensive and it becomes even unrealistic to gain well-labeled data for all target animal classes when considering the diversity. 
    
    The rise of deep neural models~\cite{resnet,LeCunCNN} brings data hunger to develop a customized high-powered model on multiple tasks. Data hunger thus becomes common when trying to train a fully supervised model. To tackle this problem, many techniques are proposed~\cite{fewshot, zeroshot1, zeroshot2}. Because, commonly, different datasets share similar feature distribution, especially when their data is sampled from close domains. To leverage such cross-domain shared knowledge, domain adaptation~\cite{hoffmantransfer,ganin2015unsupervised} has been widely studied on different tasks, such as detection~\cite{detectionadaptation, cyclegandetection}, classification~\cite{classificationadaptation1, classficationadaptation2, finegrainedadaptation, clssegdomain}, segmentation~\cite{segdomain1, segdomain2, clssegdomain} and pose estimation~\cite{posedomain3, posedomain4}. But in previous works about keypoint detection or pose estimation~\cite{posedomain1, posedomain3, xiaogangpose, bonepose}, source domain and target domain face much slighter domain shift than when transferring from human dataset to animals or among different animal species. Besides, some extra information might be available for easier knowledge transfer, such as view consistency~\cite{posedomain3}, attribute attached to samples~\cite{finegrainedadaptation} or morphological similarity~\cite{xiaogangpose, bonepose}.
    
    Domain adaptation becomes very difficult when domains face severe domain shift and no extra information is available to align feature representation on different domains, just as faced when adopting domain adaptation to animal pose estimation. A key idea in similar cases\cite{fang2018weakly, residualtransfer, hoffmantransfer} is to extract and leverage more cross-domain common features to help the final task. To reach this goal, some works~\cite{residualtransfer, hoffmantransfer} use weight-shared modules for cross-domain feature extraction. And extracted features are aligned~\cite{residualtransfer} to be represented with a more similar distribution. Besides, adversarial networks~\cite{advdomain1, advdomain2, advdomain3} or simply an adversarial loss~\cite{hoffmantransfer, finegrainedadaptation} are also used to confuse networks to focus more on domain-invariant features. In addition to the improvement of model design, data augmentation on the target domain also attracts much attention for domain adaptation. From this perspective, GAN~\cite{GAN, CycleGAN2017, instagan, constrastgan} raises many interesting temples~\cite{cyclegandetection, auggan}. But existing works still only tackle easier tasks such as object detection and when domain shift can not be well imitated by style transfer, GANs is less helpful for data augmentation. On the other hand, some works also use 'pseudo-label' for data augmentation~\cite{cyclegandetection, segdomain1, pseudolabel}. In these works, confident enough prediction on target domain data is regarded as 'pseudo ground truth' and put into training. For these works, how to select and use pseudo labels in training is critical and some special learning strategies are sometimes designed for it~\cite{segdomain1, selfpacedlearning, selfpacedcurlearning}.
    
    The aforementioned schemes show shortcomings for domain adaptation in animal pose estimation. Compared with object detection~\cite{cyclegandetection} or classification~\cite{finegrainedadaptation}, pose estimation is much more complicated and variance for pose estimation of different animals is more than texture or style difference. To this end, we propose a novel method for our task, where some popular ideas are also put into use after improvement. 
    
    \section{Preliminaries}
    \subsection{Animal Pose Dataset}
    \label{animaldataset}
\begin{figure}
        \centering
        \includegraphics[width=\linewidth]{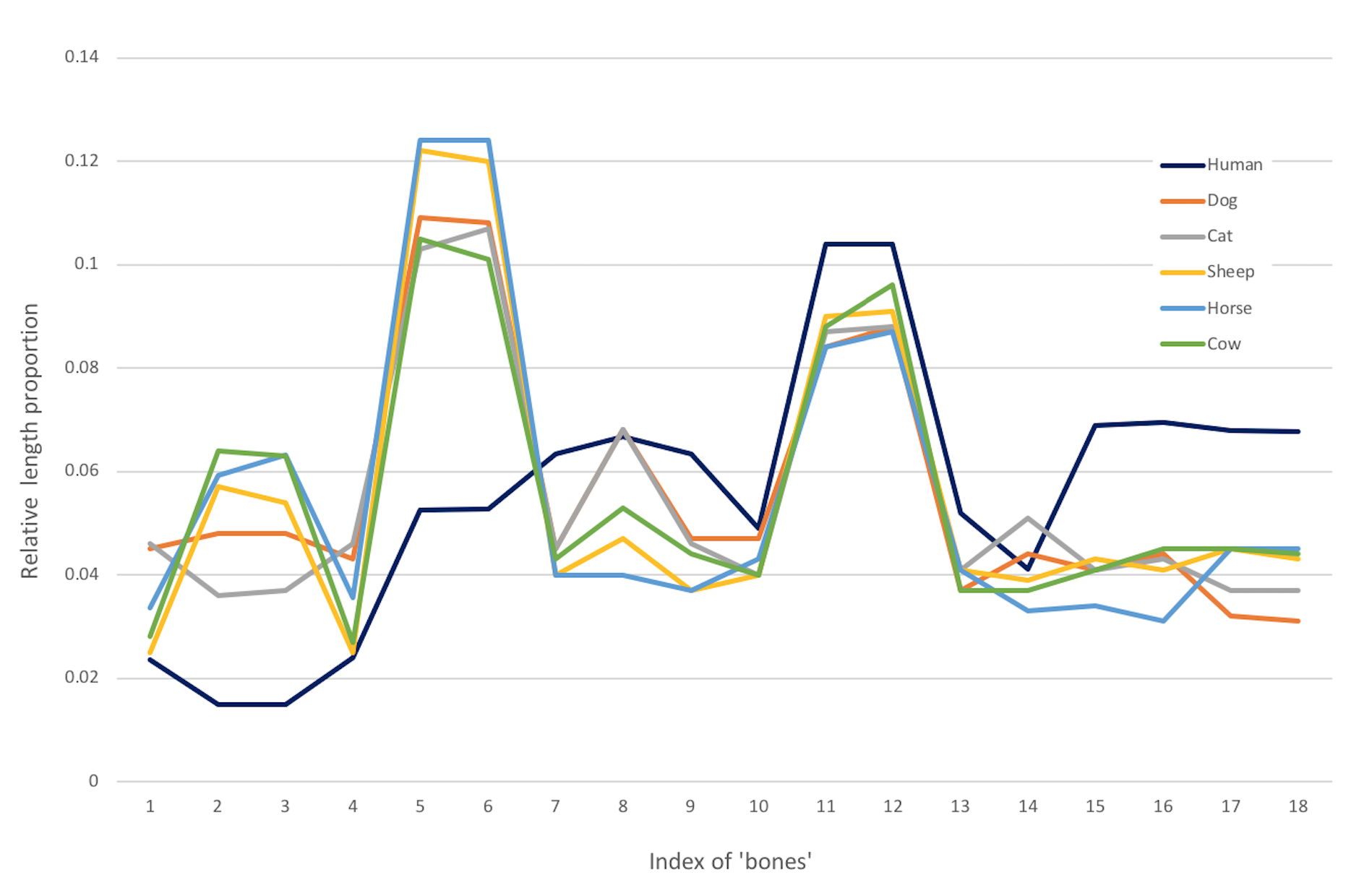}
        \caption{The length proportion of each defined ``bones'' for different classes.}
        \label{bone_proportion}
    \end{figure}

    As there are few available pose-labeled animal datasets, in order to objectively evaluate the performance on animal pose estimation and to gain basic knowledge under weak supervision, we build a pose-labeled animal dataset. Luckily, a dataset~\cite{animalvoc2011,poselets} of pose-labeled instances from VOC2011~\cite{pascal-voc-2011} is publicly available. We extend its annotation on five selected mammals: dog, cat, horse, sheep, cow. It helps to align annotation format with a popular human-keypoint format for better leveraging knowledge from human data. In this dataset, 5,517 instances of these 5 categories are distributed in more than 3,000 images. After annotation expanding, at most 20 keypoints are available on animal instances, including \emph{4 Paws, 2 Eyes, 2 Ears, 4 Elbows, Nose, Throat, Withers and Tailbase, and the 4 knees points labeled by us}. Such animal pose annotation can be aligned to that defined in popular COCO~\cite{COCO} dataset by selecting within 17 keypoints. Some dataset samples are shown in Fig \ref{animalsample}. To build such a novel dataset, only very slight labor work is involved.
    Domain shift between animals' pose and humans' pose comes mostly from the difference of their skeleton configuration, which can't be imitated by style transfer as the texture difference. We define 18 ``bones'' (link of two adjacent joints) to help explanation on it as same as those in COCO dataset. We calculate the relative length proportion of ``bones'' on average of different classes. Results are shown in Fig \ref{bone_proportion}. Some different classes of animals suffer from much slighter skeleton discrepancy than animals and humans do, which reflects the severity of domain shift different domains suffering from.  
        
    \subsection{Problem Statement}
    \label{problemstatement}
    In this paper, we aim to estimate pose configuration of animals, especially four-legged mammals. With large-scale human pose datasets and a handful of labeled animal samples available, the problem is translated into a domain adaption problem that we estimate pose on unseen animals with the help of knowledge from pose-labeled domains. This problem is formulated precisely as below.
    
    A pose-labeled dataset is denoted as $\bar{\mathcal{D}}$ consisting of both human images and mammal images: 
    
    \begin{equation}
        \bar{\mathcal{D}}=\{\bar{\mathcal{D}}_H\} \cup  \{ \bar{\mathcal{D}}_{A_i} | 1 \leq i \leq m \}
    \end{equation}
     where $m$ animal species are contained and human dataset $\bar{\mathcal{D}}_H$ is much larger than animal datasets $\bar{\mathcal{D}}_{A}$. 
     
     Each instance $\bar{I} \in \bar{\mathcal{D}}$  possesses a pose ground-truth $Y(\bar{I}) \in \mathbb{R}^{d\times2}$, which is a matrix containing ordered keypoint coordinates. Our goal is to predict underlying keypoints of unlabeled animal samples $I \in \mathcal{D}$. Their latent pose ground truth is denoted as $\hat{Y}(I)$ and is expected to be described in a uniform format with $Y(\bar{I})$. Therefore, we formulate our task as to train a model:
    
    \begin{equation}
        G_{\theta}: \mathbb{R}^{H\times W} \longrightarrow \mathbb{R}^{d\times2}
    \end{equation}
    $G_\theta$ takes an image of unseen animal species as input and predicts keypoints on it. Since prior knowledge is gained from both human data or labeled animal species, which have obvious domain shift with those unlabeled animal species. This task can thus be summarized as a cross-domain adaptation for animal pose estimation.
    
    \begin{figure*}
        \centering
        \includegraphics[width=16.6cm]{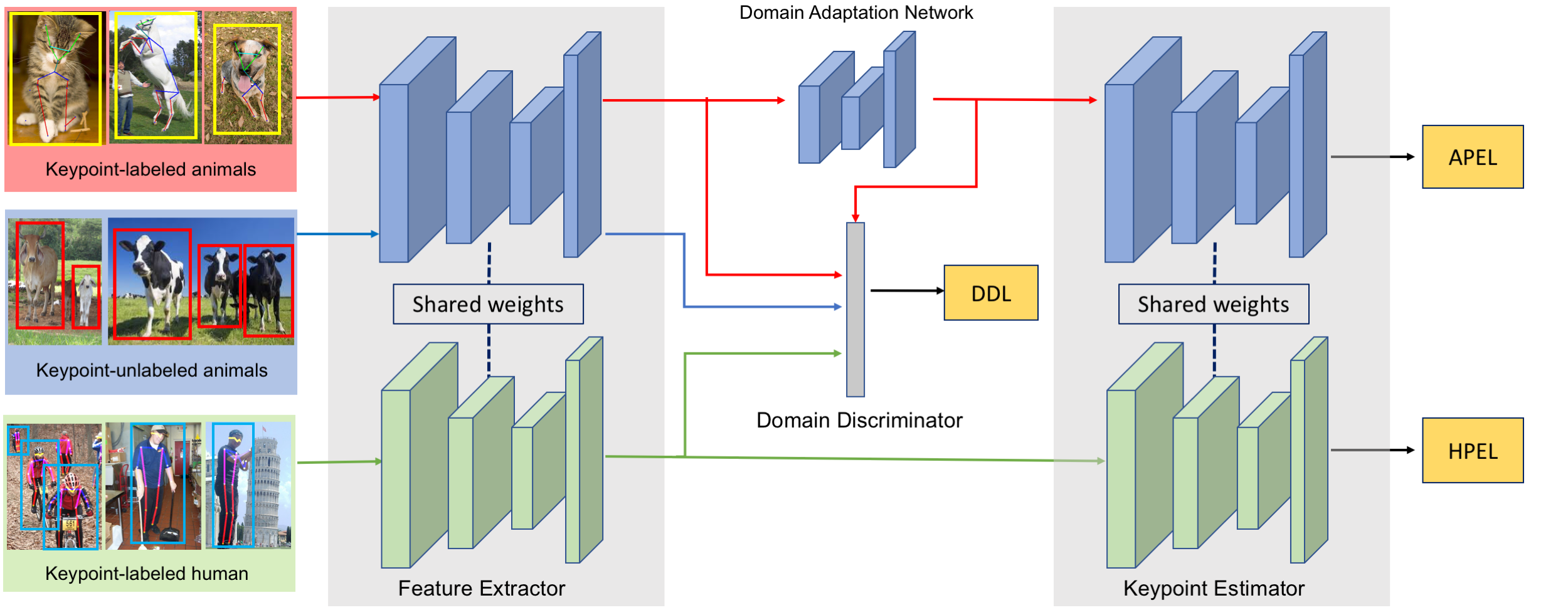}
        \caption{Pipelines in WS-CDA. Lines with color describe the flow of features along different paths. ``DDL'' indicates the domain discrimination loss. ``APEL'' and ``HPEL'' indicate animal/human pose estimation loss respectively. The cooperation of keypoint estimator and domain discriminator does not just improves the pose estimation capacity on pose-labeled samples but also forces the model to gain this through better extracting and leveraging common features shared by pose-labeled and pose-unlabeled samples.}
        \label{domain_rebalanced}
    \end{figure*}
    
    \section{Proposed Architecture}
    \label{architecture}
    Knowledge from both human dataset and animal dataset is helpful to estimate animal pose, but there exists a data imbalance problem: pose-labeled animal dataset is smaller but has slighter domain shift with the target domain while the pose-labeled human dataset is much larger but suffers from more severe domain shift. In Section \ref{WSCDA}, we design a ``Weakly- and Semi- Supervised Cross-domain Adaptation''(WS-CDA) scheme to alleviate such flaw and to better learn cross-domain shared features. In Section \ref{PPLO}, we introduce designed ``Progressive Pseudo-Label-based Optimization'' (PPLO) strategy to boost model performance on target domain referring to `pseudo-labels' for data augmentation. The final model is pre-trained through WS-CDA and boosted under PPLO.
    
    \subsection{Weakly- and Semi- supervised cross-domain adaptation(WS-CDA)}
    \label{WSCDA}
    If a model can learn more cross-domain shared features, it's reasonable to expect it to perform more robustly when facing domain shift. But single-domain data usually leads the model to learn more domain-specific and untransferable features. Based on such observations, we design WS-CDA to leverage as strong as possible cross-domain shared features for pose estimation on unseen classes. 
    
    \paragraph{Network Design} 
    As shown in Fig \ref{domain_rebalanced}, there are three sources of input data. The first is the large-scale pose-labeled human dataset, the second is a smaller pose-labeled animal dataset and the last is pose-unlabeled animal samples of an unseen class. This design uses semi-supervision because few animal samples are annotated, and weak-supervision because a large part of animal data is only labeled at a lower level (only bounding boxes are labeled). 
    
    There are four modules used in WS-CDA: 1) All data is first fed into a CNN-based module called \emph{feature extractor} to generate feature maps; 2) All feature maps would go into a \emph{domain discriminator} which distinguishes the input feature maps generated from which domain; 3) Feature maps from pose-labeled samples are also forwarded to a \emph{keypoint estimator} for supervised learning of pose estimation; 4) a \emph{domain adaptation network} is inserted to help convert the feature maps extracted to be more well represented for following pose estimation on animal instances.
    
    The losses of \emph{domain discriminator} and \emph{keypoint estimator} are set to be adversarial. As pose estimation is the main task, the \emph{domain discriminator} serves for domain confusion during feature extraction. Through this design, the model is expected to perform better on pose-unlabeled samples by leveraging better features that are shared on domains. 
    
    \paragraph{Loss Functions} 
    
    The domain discrimination loss(DDL) is defined based on cross-entropy loss as:
    \begin{equation}
    \label{DDL}
        \begin{split}
        \mathcal{L}_{DDL} = 
        & - w_1\sum^{N}_{i=1}(y_i log(\hat{y}_i) + (1 - y_i)log(1 - \hat{y}_i))\\
        & - \sum^{N}_{i=1}y_i(z_i log(\hat{z}_i) + (1-z_i)log(1-\hat{z}_i)),
        \end{split}
    \end{equation}
    where $y_i$ indicates whether $x_i$ is a human/animal sample($y_i=1$ for animals and $y_i=0$ for human); $z_i$ indicates whether $x_i$ comes from the target domain ($z_i=1$ if it is pose-unlabeled sample and otherwise $z_i=0$). $\hat{y}_i$ and $\hat{z}_i$ are predictions by the domain discriminator. $w_1$ is a weighting factor.
    
    Pose-labeled animal and human samples boost the keypoint estimator together under supervision, yielding the  ``Animal Pose Estimation Loss'' (APEL) and ``Human Pose Estimation Loss''(HPEL). The overall loss for pose estimation is as follows,
    
    \begin{equation}
        \label{pose_loss}
            \mathcal{L}_{pose} = \sum^N_{i=1}( w_2 y_i  \mathcal{L}_{A}(I_i)
            + (1-y_i) \mathcal{L}_{H}(I_i) ),
    \end{equation}
    where $\mathcal{L}_{H}$ and $\mathcal{L}_{A}$ indicate loss function of HPEL and APEL respectively and are usually both mean-square error. $w_2$ is weighting factor to alleviate the effect of dataset volume gap. Considering much more pose-labeled human samples are put into training than animal samples, without $w_2 > 1 $, model tends to perform almost equivalent to only trained on human samples.
    
    Integrated optimization target of  the framework is thus formulated as:
    \begin{equation}
        \mathcal{L}_{WS-CDA} = \alpha \mathcal{L}_{DDL} + \beta\mathcal{L}_{pose},
        \label{rcfeloss}
    \end{equation}
    with $\alpha \beta < 0$, domain discriminator and keypoint estimator are optimized adversarially, encouraging domain confusion and boosting pose estimation performance at the same time.
    
    \subsection{Progressive Pseudo-label-based Optimization (PPLO)}
    \label{PPLO}
    
    In this section, we discuss strategies designed to leverage pose-unlabeled animal samples to further boost model performance. The intuition is to approximate the underlying labels starting from the rough estimation by a 'basically-reliable' model and to select predictions on target domain with high confidence for training. These predictions are called ``pseudo-labels'' as introduced in~\cite{pseudolabel, cyclegandetection}. Considering the reliability degree of the model in different stages, we introduce an optimization method in a self-paced and alternating style for training with pseudo-labels involved. These innovations are concluded as ``Progressive Pseudo-label-based Optimization'' (PPLO) for convenience.
    
    \subsubsection{Joint learning for domain adaptation}
    In transfer learning practice, given the ground truth on both domains, adaptation can be performed in a jointly supervised scheme, which is formulated as:

    \begin{equation}
        \begin{split}
            \mathcal{L}_{joint} &= \mathcal{L}_{source} + \mathcal{L}_{target}\\
            &= \sum^{S}_{i=1}\mathcal{L}_S(I^S_i, G^S_{I_i}) +
            \sum^T_{j=1}\mathcal{L}_T(I^T_j, G^T_{I_j})
        \end{split}
        \label{jointtrain}
    \end{equation}
    where $\mathcal{L}_S$ and $\mathcal{L}_T$ are loss functions for training data respectively from source/target domain. $I^T_j$ and $I^S_i$ are samples from source/target domain whose ground truth labels are respectively $G_{I^S_i}$ and $G_{I^T_j}$.
    
    However, for unsupervised domain adaptation, this process gets stuck because no ground truth label is available on the target domain. As a sub-optimal choice, ``pseudo-label'' is introduced to fill the vacancy of it. The loss on target domain in Eq \ref{jointtrain} is transformed into:
    
    \begin{equation}
        \label{jointloss}
        \mathcal{L}_{target} = \sum^T_{j=1}\mathcal{L}_T(I^T_j, P^T_{I_j})
    \end{equation}
    where ground truth label is replaced by the selected pseudo label $P^T_{I_j}$.
    
    \subsubsection{Self-paced selection of pseudo labels}
    One main challenge to involve pseudo labels in training is that the correctness of pseudo-labels cannot be guaranteed. Instead of providing more useful knowledge of the target domain, unreliable pseudo labels will mislead model to perform worse on the target domain. To overcome this flaw, we propose a self-paced~\cite{selfpacedlearning, selfpacedcurlearning} strategy to select pseudo labels into training from easier cases into harder ones. This avoids model degradation due to aggressive use of pseudo labels. Overall, the current model prediction would be used for training as a pseudo label only when its confidence is high enough. This updates Eq \ref{jointloss} as:
    
    \begin{equation}
        \mathcal{L}^\prime_{target} = -
        \sum^T_{j=1} \hat{Y}^{(\phi)}_{j} \mathcal{L}_T(I^T_j, \mathbf{m}(I^T_j | \phi))
        \label{pploloss}
    \end{equation}
    where $\mathbf{m}(I^T_j | \phi)$ is output by model of current weights $\phi$ on $I^T_j$. $\hat{Y}^{(\phi)}_{j}$ denotes whether the pose prediction on $I^T_j$ is reliable enough:
    
    \begin{equation}
        \hat{Y}^{(\phi)}_{j} = 
        \left \{
            \begin{array}{lr}
                1, \mathbf{if} \ \mathcal{C}(\mathbf{m}(I^T_j | \phi)) > \mu & \\
                0, otherwise &
            \end{array}
        \right.
    \label{confidencefilter}
    \end{equation} 
    where $\mathcal{C}(\mathbf{m}(I^T_j | \phi))$ denotes the output confidence score on $I^T_j$ by the current model. $\mu$ is the threshold to filter unreliable outputs. In the design of self-paced selection, restrict of $\mu$ keeps being relaxed during the optimization of model.
    
    \subsubsection{Alternating cross-domain training}
    \label{alternative}
    WS-CDA and the careful self-paced selection of pseudo labels make pseudo labels involved in training already much more reliable. However, pseudo labels still contain more noise than real ground truth, bringing a risk of model degradation. To relieve the effect, the model is jointly trained in a cautious manner where samples from source domains and target domains are fed into training alternately. 
    
    If the data volume of the source domain and target domain is close, such alternating training approximates to training on the mixture of data from both domains. However, domains suffer from huge data volume gap in our task where pose-labeled animal samples are much more than target domain samples with pseudo labels. In such a case, training on a mixed dataset will lead the model to learn more from the domain with more samples while alternating training relieves the problem.
    
\begin{figure}
        \centering
        \includegraphics[width=\linewidth]{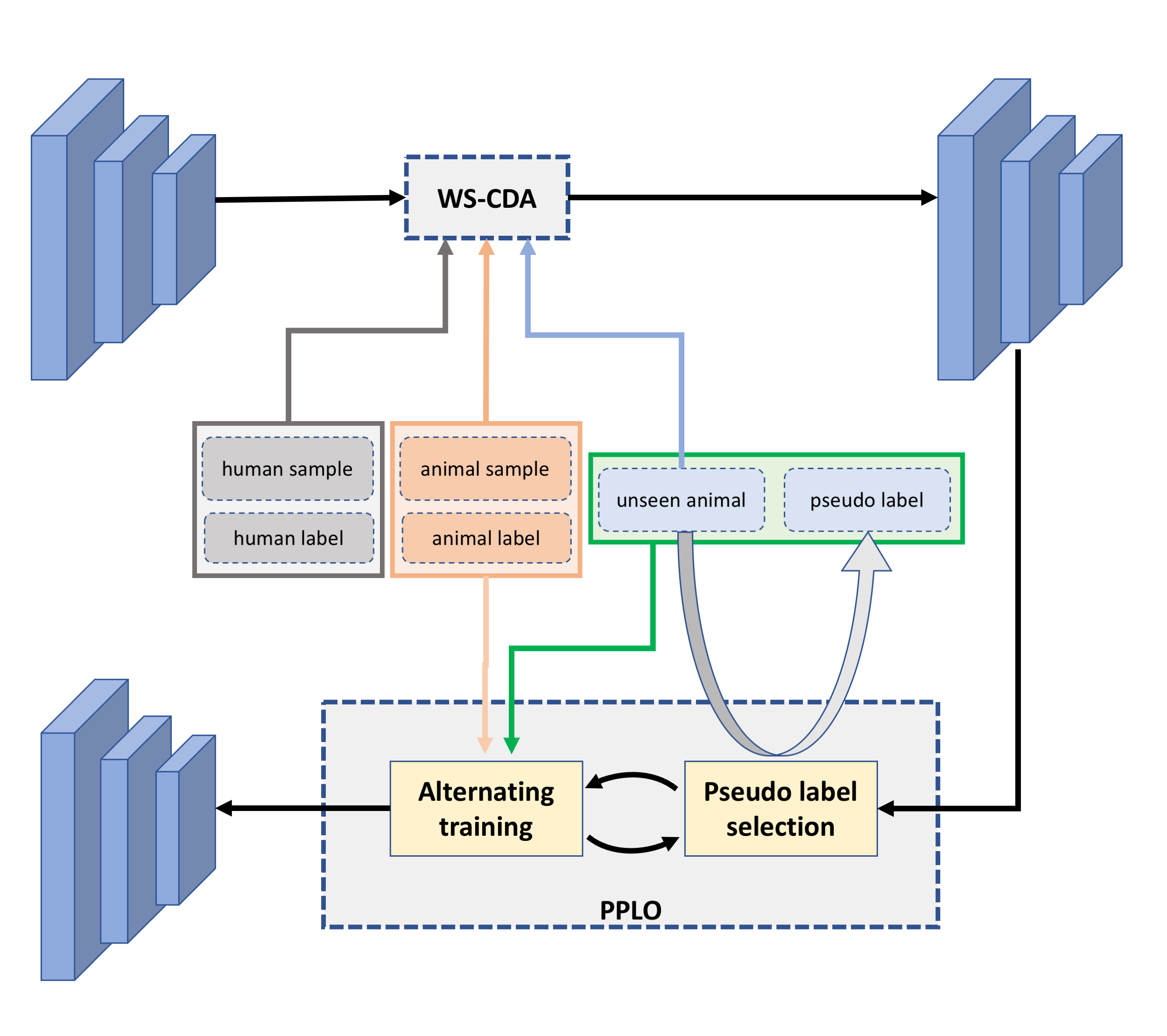}
        \caption{The overall process of our proposed scheme with WS-CDA and PPLO both involved. Blue blocks indicate the overall network in Fig \ref{domain_rebalanced}}
        \label{overall_process}
    \end{figure}

    The procedure of PPLO in one epoch is explained in Algorithm \ref{pploalg}. The overall design of our proposed scheme for domain adaptation based on multiple domain data is illustrated as Fig \ref{overall_process}.
    
    \begin{algorithm}
    \caption{overall procedure of PPLO} 
    \label{pploalg} 
    \begin{algorithmic}[1] 
    \REQUIRE ~~\\ 
    1. Current model weights, $\phi$\\
    2. Current threshold to filter unreliable pseudo-label, $\mu$\\
    3. Source domain data $\bar{I} \in \bar{\mathcal{D}}$\\
    4. Ground truth on source domain $G^{S}_{\bar{I}} \in G^{S}$\\
    5. Target domain data $I \in \mathcal{D}$\\
    6. Pseudo labels on target domain $P^{T}_I \in P^T$\\
    7. Training steps on source domain $K_S$\\
    8. Training steps on target domain $K_T$\\
    9. Strategy to relax value of $\mu$, $\mathcal{S}$\\
    \ENSURE ~~\\ 
    1. Updated model weights $\phi$\\
    2. Updated value of $\mu$\\
    3. Updated set of $P^T$\\
    \FOR{t = 1,...,$K_S$}
        \STATE Sample a mini-batch $\mathcal{B}_{\bar{I}}$ from $\bar{\mathcal{D}}$.
        \STATE Update $\phi$ by training on $\mathcal{B}_{\bar{I}}$ with $G^S_{\bar{I}}$.
    \ENDFOR
    \FOR{each $I \in D$}
        \STATE Predict keypoints $\mathcal{K}_{I}$ of $I$.
        \IF{confidence of $\mathcal{K}_{I}$ $> \mu $}
            \STATE update the pseudo label of $I$ in $P^T$ to be $\mathcal{K}_{I}$
        \ENDIF
    \ENDFOR
    \FOR{t = 1,...,$K_T$}
        \STATE Sample a mini-batch $\mathcal{B}_I$ from $ {\mathcal{D}}$ with $P^T_I \in P^T$.
        \STATE Update $\phi$ by training on $\mathcal{B}_I$ with $P^T_I$.
    \ENDFOR
    \STATE update $\mu$ with given strategy $\mathcal{S}$.
    \RETURN $\phi$, $\mu$, $P^T$; 
    \end{algorithmic}
    \end{algorithm}
    
    \section{Evaluation}
    We evaluate the effectiveness of the proposed designs in this section. Because there are few existing methods available for animal pose estimation without labeling too many samples, we try to build comparisons by transplanting some previous works focusing on similar tasks.
    
    \subsection{Experiment Settings}
    We transplant some popular multiple pose estimation frameworks~\cite{alphapose, cpn, maskrcnn} to do animal pose estimation for comparison. Furthermore, we also compare different popular domain adaptation methods~\cite{hoffmantransfer,residualtransfer,cyclegandetection}.
    
    For fairness, the data sources are limited in experiments. The pose-labeled human dataset for training is the full COCO2017 train dataset~\cite{COCO}, which contains 100k+ instances, much larger than the built pose-labeled animal dataset. Our built dataset is the only source of pose-labeled animal samples and pose-unlabeled animal samples come from the COCO2017 train dataset~\cite{COCO} of the detection task. 
    
    Unless otherwise specified, all models are realized as defined in officially released code by default. But we explain in detail the configuration of adopted ``AlphaPose'' model: i) \emph{feature extractor} and \emph{domain adaptation networks}(DAN) are both based on ResNet-101~\cite{resnet}. ii) a SE module~\cite{senet} is inserted between neighboring residual blocks; iii) \emph{keypoint estimator} consists of two DUC~\cite{duc} layers. The model outputs a heatmap for each keypoint with a confidence ( $\mathcal{C}(\cdot)$ in Eq \ref{confidencefilter}) to filter unreliable detected keypoint candidates.
    
    The training procedure is also standardized: AlphaPose-based models are all trained in 3 steps: i) training with learning rate=$1e-4$; ii) training with learning rate=$1e-4$ and disturbance (noise and dithering of patch cropping) added; iii) training with learning rate=$1e-5$ and disturbance added. Model is optimized by RMSprop~\cite{optimizers} with default parameters in Pytorch~\cite{pytorch}. Training goes to the next stage or ends when the loss converges stably. 
    
    Lastly, unless otherwise specified, hyperparameters are necessarily uniform. For WS-CDA parameters, we set $\alpha=-1$, $\beta=500$, $w_1=1$ and $w_2=10$. We set the initial value of $\mu$ to be 0.9 in Algorithm \ref{pploalg} and it decreases by 0.01 after every 10 epochs if some pseudo label is updated during the recent 10 epochs. Training batch size is always set to be 64.
    
    \begin{figure*}[!htbp]
        \centering
        \includegraphics[width=17cm]{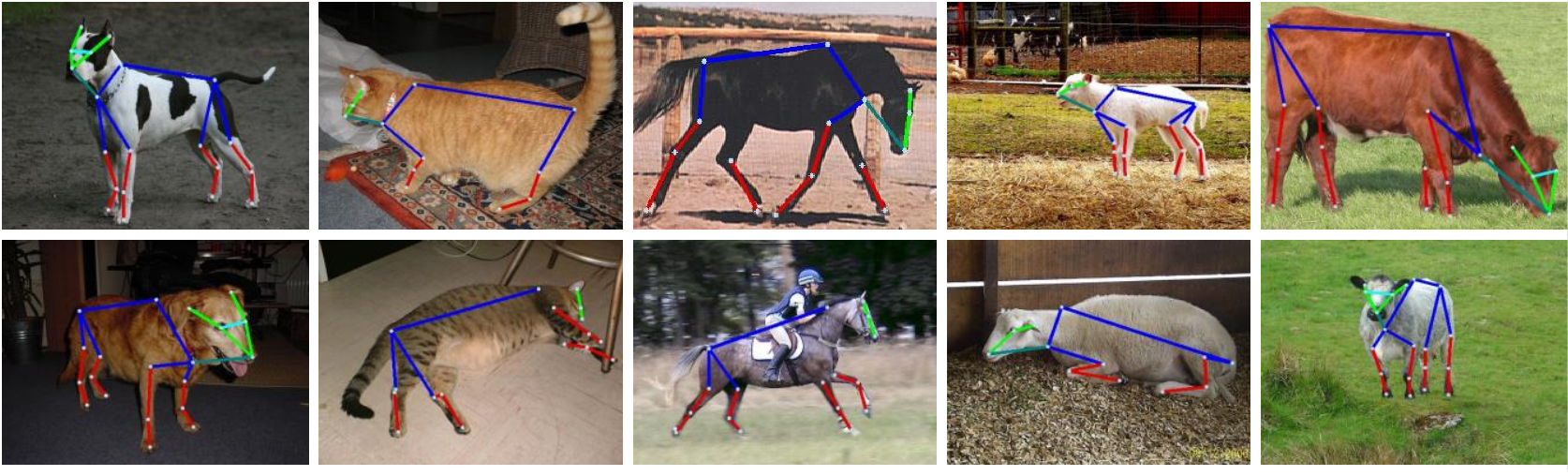}
        \caption{cross-domain adaptation results by our proposed scheme on unseen animals included in our built dataset.}
        \label{pplo_samples}
    \end{figure*}
    
    \begin{figure*}[!htbp]
        \centering
        \includegraphics[width=17cm]{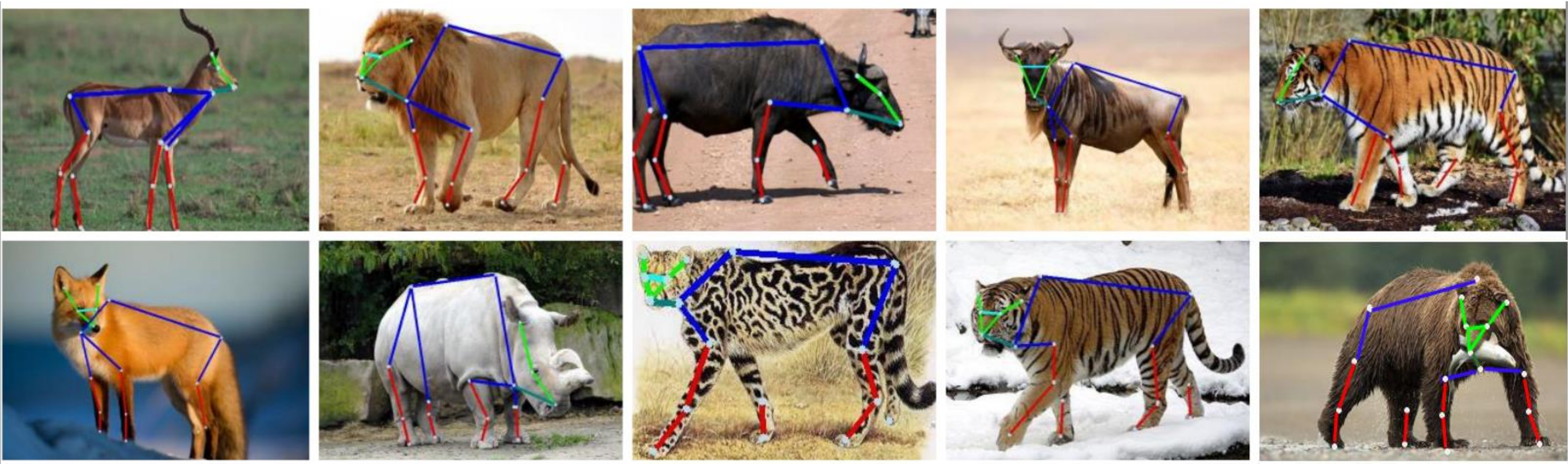}
        \caption{cross-domain adaptation results by our proposed scheme on unseen animals not included in our built dataset.}
        \label{pplo_samples_others}
    \end{figure*}
    
    \subsection{Evaluation for WS-CDA}
    To evaluate the effectiveness of WS-CDA precisely, we set experiment groups with different modules enabled or with different training data used and all groups use the 'AlphaPose' framework as described before. Details are reported in Table \ref{wscda_results}. We select 1,117 instances from the built animal pose dataset to for testing. 

    \begin{table}
    \begin{center}
    \begin{tabular}{c|ccccc|c}
    \toprule
    Index & $N_A$ & H & DAN & UA  & RB & mAP \\
    \hline
    1 & 0  & $\checkmark$ &  &  &  & 0.4 \\
    \hline
    2 & 2k & & & & & 30.3 \\
    3 & 2k & \checkmark &  &  & & 51.5 \\
    4 & 2k & \checkmark & \checkmark &  &  & 53.0 \\
    5 & 2k & \checkmark & \checkmark & \checkmark & & 45.7\\
    6 & 2k & \checkmark & \checkmark & \checkmark & \checkmark  & \textbf{56.7}  \\
    \hline
    7 & 4k &  &  &  &  & 44.3 \\
    8 & 4k & \checkmark & &  &  & 62.3 \\
    9 & 4k & \checkmark & \checkmark &  & & 63.1 \\
    10 & 4k & \checkmark & \checkmark & \checkmark &  & 57.2 \\
    11 & 4K & \checkmark & \checkmark & \checkmark & \checkmark & \textbf{65.7} \\
    \bottomrule
    \end{tabular}
    \end{center}
    \caption{Evaluation result of WS-CDA under different settings. mAP is calculated with COCO-api~\cite{cocoapi}. $N_A$ is the number of pose-labeled animal instances used for training. H indicates whether human data is used for training. DAN indicates whether the domain adaptation network is used. UA indicates whether pose-unlabeled animal data is used. $w_2 = 10$ if RB is enabled, otherwise $w_2=1$.}
    \label{wscda_results}
    \end{table}
    
    Experiment results show that when only trained on the human dataset, even if it is larger and well-labeled, the model encounters total failure on the animal test set. After a handful of pose-labeled animal samples have been added to training, the model performance leaps. Such a difference obviously comes from the huge domain shift between animals and humans. Furthermore, even though training on solely human data fails dramatically, adding the same set of human samples into training together with animal samples, it boosts model performance significantly. It proves that there still be many common features that help pose estimation on both humans and animals, while training solely on human data misleads model to more human-specific features instead of them. Then, experiments suggest that domain adaptation network and the weighting factor alleviate the negative influence of the volume gap between animal dataset and human dataset. And the pose-unlabeled animal samples would help when the weighting factor is enabled, otherwise, it might instead degrade the model.
    
    \subsection{Evaluation on unseen species}
    
    \begin{table}
        \begin{center}
        \tabcolsep=2.0pt
        \begin{tabular}{@{}lccccc}
        \toprule
        & \multicolumn{5}{c}{mAP for each class}\\
        \cline{2-6}
        Method & cat & dog & sheep & cow & horse\\
        \hline
        baseline & 16.9 & 17.2 & 38.3 & 35.5 & 28.9\\
        \hline
        \textsl{w/o adaptation}\\
        \lvl Maskrcnn~\cite{maskrcnn} & 22.5 & 21.6 & 18.7 & 21.6 & 23.6\\
        \lvl AlphaPose~\cite{alphapose} & 37.6 & 37.3 & 49.4 & 50.3 & 47.9\\
        \lvl CPN~\cite{cpn} & 30.7 & 37.8 & 51.1 & 51.2 & 41.2\\
        \hline
        \textsl{w/ adaptation}\\
        \lvl CycleGAN+PL~\cite{cyclegandetection} & 35.9 & 36.7 & 48.0 & 50.1 & 48.1\\
        \lvl dom confusion~\cite{hoffmantransfer} & 38.0 & 37.7 & 49.5 & 50.6 & 48.5\\ 
        \lvl residual transfer~\cite{residualtransfer} & 37.8 & 38.2 & 49.1 & 50.8 & 48.6\\
        \hline
        \textsl{proposed}\\
        \lvl WS-CDA ($w_2=1$) & 34.5 & 32.3 & 47.6 & 47.8 & 46.2\\
        \lvl WS-CDA & 39.2 & 38.6 & 51.3 & 54.6 & 50.3\\
        \lvl PPLO & 37.9 & 37.5 & 49.3 & 50.3 & 48.1 \\
        \lvl WS-CDA+PPLO & \textbf{42.3} & \textbf{41.0} & \textbf{54.7} & \textbf{57.3} & \textbf{53.1}\\
        \bottomrule
        \end{tabular}
        \end{center}
        \caption{Comparisons in term of mAP using different pose estimation frameworks and with/without domain adaptation to the target unseen animal class.}
        \label{animaltoanimal}
    \end{table}
    
    We design experiments to evaluate the performance of pose estimation on an unseen animal class. ``Unseen'' means the animal samples for test come from classes/domains not involved in training. Among the five pose-labeled animal classes, we simply set one class as the test set and the other four for training. In Table \ref{animaltoanimal}, the baseline model is trained solely on the animal dataset. For other groups, models are trained with pose-labeled human dataset involved. For the ``w/o adaptation'' group, models are pre-triained on the human dataset and then simply fine-tuned on animal samples. We bring some other domain adaptation methods~\cite{cyclegandetection, hoffmantransfer, residualtransfer} into evaluation for comparison. For method in~\cite{CycleGAN2017}, Cyclegan is used for data augmentation with extra animal samples used~\cite{catdogdataset}. The method in~\cite{hoffmantransfer} is adopted without the ``softlabel loss'' involved. For method in~\cite{residualtransfer}, we use residual transfer networks based on fully connected networks to replace the adversarial domain discriminator.

    The experiment proves the effectiveness of human prior knowledge, WS-CDA, and PPLO when performing cross-domain adaptation for pose estimation on unseen animal classes. Furthermore, our proposed method outperforms other domain adaptation techniques. An interesting fact is that GAN-based methods hardly show good effectiveness, even though they achieve impressive performance in some other tasks~\cite{cyclegandetection}. We conclude it to the failure of the original pose label when augmenting training data with GANs~\cite{CycleGAN2017, instagan, constrastgan}. To be precise, GANs only leave segmentation mask unchanged after transformation but joint locations are usually changed, which invalidates the original pose label. Such data augmentation introduces much noise of labels into training and probably leads to model degradation. 
    
    \section{Conclusion}
    We proposed a novel task of pose estimation on unseen animals with domain adaptation. A novel cross-domain adaptation mechanism is developed for this task. We designed a ``weakly- and semi-supervised cross-domain adaptation'' (WS-CDA) scheme to transfer knowledge from human and animal data to unseen animals. Furthermore, we designed a ``progressive pseudo-label-based optimization''(PPLO) to boost model performance by bringing target domain data into training with `pseudo-labels', for which a self-paced `pseudo-label' selection method and an alternating training method are introduced. To facilitate similar future tasks, we built an animal pose dataset providing novel prior knowledge. Experiments prove the effectiveness of our proposed scheme, which achieves human-level pose estimation accuracy on animal pose estimation. 
    
    \section*{Acknowledgement}
    This work is supported in part by the National Key R\&D Program of China, No.2017YFA0700800, National Natural Science Foundation of China under Grants 61772332. We also thank all annotators contributing to this work. We acknowledge the support from many annotators to build the dataset, including some of our friends and relatives. I also acknowledge Gary Yang and Ilian Herzi for their helpful suggestions on paper writing.

    {\small
    \bibliographystyle{ieee_fullname}
    \bibliography{egbib}

\begin{thebibliography}{10}\itemsep=-1pt

\bibitem{MPII}
Mykhaylo Andriluka, Leonid Pishchulin, Peter Gehler, and Bernt Schiele.
\newblock 2d human pose estimation: New benchmark and state of the art
  analysis.
\newblock In {\em CVPR}, pages 3686--3693, 2014.

\bibitem{animalvoc2011}
Lubomir Bourdev.
\newblock Dataset of keypoints and foreground annotations for all categories of
  pascal 2011, Feb 2012.

\bibitem{poselets}
Lubomir Bourdev and Jitendra Malik.
\newblock Poselets: Body part detectors trained using 3d human pose
  annotations.
\newblock In {\em ICCV}, pages 1365--1372, 2009.

\bibitem{advdomain3}
Konstantinos Bousmalis, Nathan Silberman, David Dohan, Dumitru Erhan, and Dilip
  Krishnan.
\newblock Unsupervised pixel-level domain adaptation with generative
  adversarial networks.
\newblock In {\em CVPR}, volume~1, page~7, 2017.

\bibitem{advdomain1}
Zhangjie Cao, Lijia Ma, Mingsheng Long, and Jianmin Wang.
\newblock Partial adversarial domain adaptation.
\newblock {\em CoRR}, abs/1808.04205, 2018.

\bibitem{cao2017realtime}
Zhe Cao, Tomas Simon, Shih-En Wei, and Yaser Sheikh.
\newblock Realtime multi-person 2d pose estimation using part affinity fields.
\newblock In {\em CVPR}, pages 7291--7299, 2017.

\bibitem{detectionadaptation}
Rita Chattopadhyay, Qian Sun, Wei Fan, Ian Davidson, Sethuraman Panchanathan,
  and Jieping Ye.
\newblock Multisource domain adaptation and its application to early detection
  of fatigue.
\newblock {\em ACM Transactions on Knowledge Discovery from Data (TKDD)}, 6:18,
  2012.

\bibitem{cpn}
Yilun Chen, Zhicheng Wang, Yuxiang Peng, Zhiqiang Zhang, Gang Yu, and Jian Sun.
\newblock Cascaded pyramid network for multi-person pose estimation.
\newblock In {\em CVPR}, pages 7103--7112, 2018.

\bibitem{posedomain1}
Chia-Jung Chou, Jui-Ting Chien, and Hwann-Tzong Chen.
\newblock Self adversarial training for human pose estimation.
\newblock {\em CoRR}, abs/1707.02439, 2017.

\bibitem{handpose1}
Ali Erol, George Bebis, Mircea Nicolescu, Richard~D Boyle, and Xander Twombly.
\newblock Vision-based hand pose estimation: A review.
\newblock {\em CVIU}, 108:52--73, 2007.

\bibitem{pascal-voc-2011}
M. Everingham, L. Van~Gool, C.~K.~I. Williams, J. Winn, and A. Zisserman.
\newblock The {PASCAL} {V}isual {O}bject {C}lasses {C}hallenge 2011 {(VOC2011)}
  {R}esults.
\newblock
  http://www.pascal-network.org/challenges/VOC/voc2011/workshop/index.html.

\bibitem{headpose2}
Gabriele Fanelli, Juergen Gall, and Luc Van~Gool.
\newblock Real time head pose estimation with random regression forests.
\newblock In {\em CVPR}, pages 617--624, 2011.

\bibitem{fang2018weakly}
Hao-Shu Fang, Guansong Lu, Xiaolin Fang, Jianwen Xie, Yu-Wing Tai, and Cewu Lu.
\newblock Weakly and semi supervised human body part parsing via pose-guided
  knowledge transfer.
\newblock In {\em CVPR}, pages 70--78. IEEE, 2018.

\bibitem{alphapose}
Hao-Shu Fang, Shuqin Xie, Yu-Wing Tai, and Cewu Lu.
\newblock {RMPE}: Regional multi-person pose estimation.
\newblock In {\em ICCV}, 2017.

\bibitem{ganin2015unsupervised}
Yaroslav Ganin and Victor Lempitsky.
\newblock Unsupervised domain adaptation by backpropagation.
\newblock In {\em ICML}, pages 1180--1189, 2015.

\bibitem{clssegdomain}
Weifeng Ge, Sibei Yang, and Yizhou Yu.
\newblock Multi-evidence filtering and fusion for multi-label classification,
  object detection and semantic segmentation based on weakly supervised
  learning.
\newblock In {\em CVPR}, pages 1277--1286, 2018.

\bibitem{finegrainedadaptation}
Timnit Gebru, Judy Hoffman, and Li Fei-Fei.
\newblock Fine-grained recognition in the wild: A multi-task domain adaptation
  approach.
\newblock In {\em ICCV}, pages 1358--1367, 2017.

\bibitem{2steppose2}
Georgia Gkioxari, Bharath Hariharan, Ross Girshick, and Jitendra Malik.
\newblock Using k-poselets for detecting people and localizing their keypoints.
\newblock In {\em CVPR}, pages 3582--3589, 2014.

\bibitem{classificationadaptation1}
Xavier Glorot, Antoine Bordes, and Yoshua Bengio.
\newblock Domain adaptation for large-scale sentiment classification: A cdeep
  learning approach.
\newblock In {\em ICML}, pages 513--520, 2011.

\bibitem{GAN}
Ian Goodfellow, Jean Pouget-Abadie, Mehdi Mirza, Bing Xu, David Warde-Farley,
  Sherjil Ozair, Aaron Courville, and Yoshua Bengio.
\newblock Generative adversarial nets.
\newblock In {\em NIPS}, pages 2672--2680, 2014.

\bibitem{classficationadaptation2}
Raghuraman Gopalan, Ruonan Li, and Rama Chellappa.
\newblock Domain adaptation for object recognition: An unsupervised approach.
\newblock In {\em ICCV}, pages 999--1006, 2011.

\bibitem{maskrcnn}
Kaiming He, Georgia Gkioxari, Piotr Doll{\'a}r, and Ross Girshick.
\newblock Mask r-cnn.
\newblock In {\em CVPR}, pages 2961--2969, 2017.

\bibitem{resnet}
Kaiming He, Xiangyu Zhang, Shaoqing Ren, and Jian Sun.
\newblock Deep residual learning for image recognition.
\newblock In {\em CVPR}, pages 770--778, 2016.

\bibitem{senet}
Jie Hu, Li Shen, and Gang Sun.
\newblock Squeeze-and-excitation networks.
\newblock 2018.

\bibitem{auggan}
Sheng-Wei Huang, Che-Tsung Lin, Shu-Ping Chen, Yen-Yi Wu, Po-Hao Hsu, and
  Shang-Hong Lai.
\newblock Auggan: Cross domain adaptation with gan-based data augmentation.
\newblock In {\em ECCV}, pages 718--731, 2018.

\bibitem{cyclegandetection}
Naoto Inoue, Ryosuke Furuta, Toshihiko Yamasaki, and Kiyoharu Aizawa.
\newblock Cross-domain weakly-supervised object detection through progressive
  domain adaptation.
\newblock In {\em CVPR}, pages 5001--5009, 2018.

\bibitem{selfpacedcurlearning}
Lu Jiang, Deyu Meng, Qian Zhao, Shiguang Shan, and Alexander~G Hauptmann.
\newblock Self-paced curriculum learning.
\newblock In {\em AAAI}, pages 2694--2700, 2015.

\bibitem{pseudolabel}
Guoliang Kang, Liang Zheng, Yan Yan, and Yi Yang.
\newblock Deep adversarial attention alignment for unsupervised domain
  adaptation: the benefit of target expectation maximization.
\newblock In {\em ECCV}, pages 401--416, 2018.

\bibitem{handpose2}
Cem Keskin, Furkan K{\i}ra{\c{c}}, Yunus~Emre Kara, and Lale Akarun.
\newblock Hand pose estimation and hand shape classification using
  multi-layered randomized decision forests.
\newblock In {\em ECCV}, pages 852--863, 2012.

\bibitem{selfpacedlearning}
M~Pawan Kumar, Benjamin Packer, and Daphne Koller.
\newblock Self-paced learning for latent variable models.
\newblock In {\em NIPS}, pages 1189--1197, 2010.

\bibitem{LeCunCNN}
Yann LeCun, Bernhard Boser, John~S Denker, Donnie Henderson, Richard~E Howard,
  Wayne Hubbard, and Lawrence~D Jackel.
\newblock Backpropagation applied to handwritten zip code recognition.
\newblock {\em Neural Computation}, 1:541--551, 1989.

\bibitem{headpose3}
Renxiang Li, Carl~M Danielsen, and Cuneyt~M Taskiran.
\newblock Apparatus and methods for head pose estimation and head gesture
  detection, Aug 2008.
\newblock US Patent 7,412,077.

\bibitem{constrastgan}
Xiaodan Liang, Hao Zhang, Liang Lin, and Eric Xing.
\newblock Generative semantic manipulation with mask-contrasting gan.
\newblock In {\em ECCV}, pages 558--573, 2018.

\bibitem{cocoapi}
Tsung-Yi Lin.
\newblock https://github.com/cocodataset/cocoapi.

\bibitem{COCO}
Tsung-Yi Lin, Michael Maire, Serge Belongie, James Hays, Pietro Perona, Deva
  Ramanan, Piotr Doll{\'a}r, and C~Lawrence Zitnick.
\newblock Microsoft coco: Common objects in context.
\newblock In {\em ECCV}, pages 740--755. Springer, 2014.

\bibitem{residualtransfer}
Mingsheng Long, Han Zhu, Jianmin Wang, and Michael~I Jordan.
\newblock Unsupervised domain adaptation with residual transfer networks.
\newblock In {\em NIPS}, pages 136--144, 2016.

\bibitem{instagan}
Sangwoo Mo, Minsu Cho, and Jinwoo Shin.
\newblock Instagan: Instance-aware image-to-image translation.
\newblock In {\em ICLR}, 2019.

\bibitem{headpose1}
Erik Murphy-Chutorian and Mohan~Manubhai Trivedi.
\newblock Head pose estimation in computer vision: A survey.
\newblock {\em IEEE Trans. Pattern Anal. Mach. Intell.}, 31:607--626, 2009.

\bibitem{catdogdataset}
O.~M. Parkhi, A. Vedaldi, A. Zisserman, and C.~V. Jawahar.
\newblock Cats and dogs.
\newblock In {\em CVPR}, 2012.

\bibitem{pytorch}
Adam Paszke, Sam Gross, Soumith Chintala, Gregory Chanan, Edward Yang, Zachary
  DeVito, Zeming Lin, Alban Desmaison, Luca Antiga, and Adam Lerer.
\newblock Automatic differentiation in pytorch.
\newblock In {\em NIPS}, 2017.

\bibitem{2steppose1}
Leonid Pishchulin, Arjun Jain, Mykhaylo Andriluka, Thorsten Thorm{\"a}hlen, and
  Bernt Schiele.
\newblock Articulated people detection and pose estimation: Reshaping the
  future.
\newblock In {\em CVPR}, pages 3178--3185, 2012.

\bibitem{animalface1}
Maheen Rashid, Xiuye Gu, and Yong Jae~Lee.
\newblock Interspecies knowledge transfer for facial keypoint detection.
\newblock In {\em CVPR}, pages 6894--6903, 2017.

\bibitem{optimizers}
Sebastian Ruder.
\newblock An overview of gradient descent optimization algorithms.
\newblock {\em CoRR}, abs/1609.04747, 2016.

\bibitem{fewshot}
Jake Snell, Kevin Swersky, and Richard Zemel.
\newblock Prototypical networks for few-shot learning.
\newblock In {\em NIPS}, pages 4077--4087, 2017.

\bibitem{zeroshot1}
Richard Socher, Milind Ganjoo, Christopher~D Manning, and Andrew Ng.
\newblock Zero-shot learning through cross-modal transfer.
\newblock In {\em NIPS}, pages 935--943, 2013.

\bibitem{posedomain4}
Adrian Spurr, Jie Song, Seonwook Park, and Otmar Hilliges.
\newblock Cross-modal deep variational hand pose estimation.
\newblock In {\em CVPR}, pages 89--98, 2018.

\bibitem{animalface3}
James Thewlis, Hakan Bilen, and Andrea Vedaldi.
\newblock Unsupervised learning of object landmarks by factorized spatial
  embeddings.
\newblock In {\em ICCV}, pages 5916--5925, 2017.

\bibitem{deeppose}
Alexander Toshev and Christian Szegedy.
\newblock Deeppose: Human pose estimation via deep neural networks.
\newblock In {\em CVPR}, pages 1653--1660, 2014.

\bibitem{hoffmantransfer}
Eric Tzeng, Judy Hoffman, Trevor Darrell, and Kate Saenko.
\newblock Simultaneous deep transfer across domains and tasks.
\newblock In {\em ICCV}, pages 4068--4076, 2015.

\bibitem{advdomain2}
Eric Tzeng, Judy Hoffman, Kate Saenko, and Trevor Darrell.
\newblock Adversarial discriminative domain adaptation.
\newblock In {\em CVPR}, volume~1, page~4, 2017.

\bibitem{duc}
Panqu Wang, Pengfei Chen, Ye Yuan, Ding Liu, Zehua Huang, Xiaodi Hou, and
  Garrison Cottrell.
\newblock Understanding convolution for semantic segmentation.
\newblock In {\em WACV}, pages 1451--1460, 2018.

\bibitem{animalface2}
Heng Yang, Renqiao Zhang, and Peter Robinson.
\newblock Human and sheep facial landmarks localisation by triplet interpolated
  features.
\newblock In {\em WACV}, pages 1--8. IEEE, 2016.

\bibitem{xiaogangpose}
Wei Yang, Wanli Ouyang, Xiaolong Wang, Jimmy Ren, Hongsheng Li, and Xiaogang
  Wang.
\newblock 3d human pose estimation in the wild by adversarial learning.
\newblock In {\em CVPR}, volume~1, 2018.

\bibitem{segdomain2}
Yang Zhang, Philip David, and Boqing Gong.
\newblock Curriculum domain adaptation for semantic segmentation of urban
  scenes.
\newblock In {\em ICCV}, volume~2, page~6, 2017.

\bibitem{zeroshot2}
Ziming Zhang and Venkatesh Saligrama.
\newblock Zero-shot learning via semantic similarity embedding.
\newblock In {\em ICCV}, pages 4166--4174, 2015.

\bibitem{bonepose}
Xingyi Zhou, Qixing Huang, Xiao Sun, Xiangyang Xue, and Yichen Wei.
\newblock Towards 3d human pose estimation in the wild: a weakly-supervised
  approach.
\newblock In {\em ICCV}, 2017.

\bibitem{posedomain3}
Xingyi Zhou, Arjun Karpur, Chuang Gan, Linjie Luo, and Qixing Huang.
\newblock Unsupervised domain adaptation for 3d keypoint estimation via view
  consistency.
\newblock In {\em ECCV}, pages 137--153, 2018.

\bibitem{CycleGAN2017}
Jun-Yan Zhu, Taesung Park, Phillip Isola, and Alexei~A Efros.
\newblock Unpaired image-to-image translation using cycle-consistent
  adversarial networks.
\newblock In {\em ICCV}, 2017.

\bibitem{segdomain1}
Yang Zou, Zhiding Yu, BVK~Vijaya Kumar, and Jinsong Wang.
\newblock Unsupervised domain adaptation for semantic segmentation via
  class-balanced self-training.
\newblock In {\em ECCV}, pages 289--305, 2018.

\bibitem{animalpose}
Silvia Zuffi, Angjoo Kanazawa, David~W Jacobs, and Michael~J Black.
\newblock 3d menagerie: Modeling the 3d shape and pose of animals.
\newblock In {\em CVPR}, pages 5524--5532, 2017.

\end{thebibliography}
    }
    \newpage
    \appendix
    \section*{Appendix}
    \section{Proposed Dataset}
\begin{figure*}
        \centering
        \includegraphics[width=17cm]{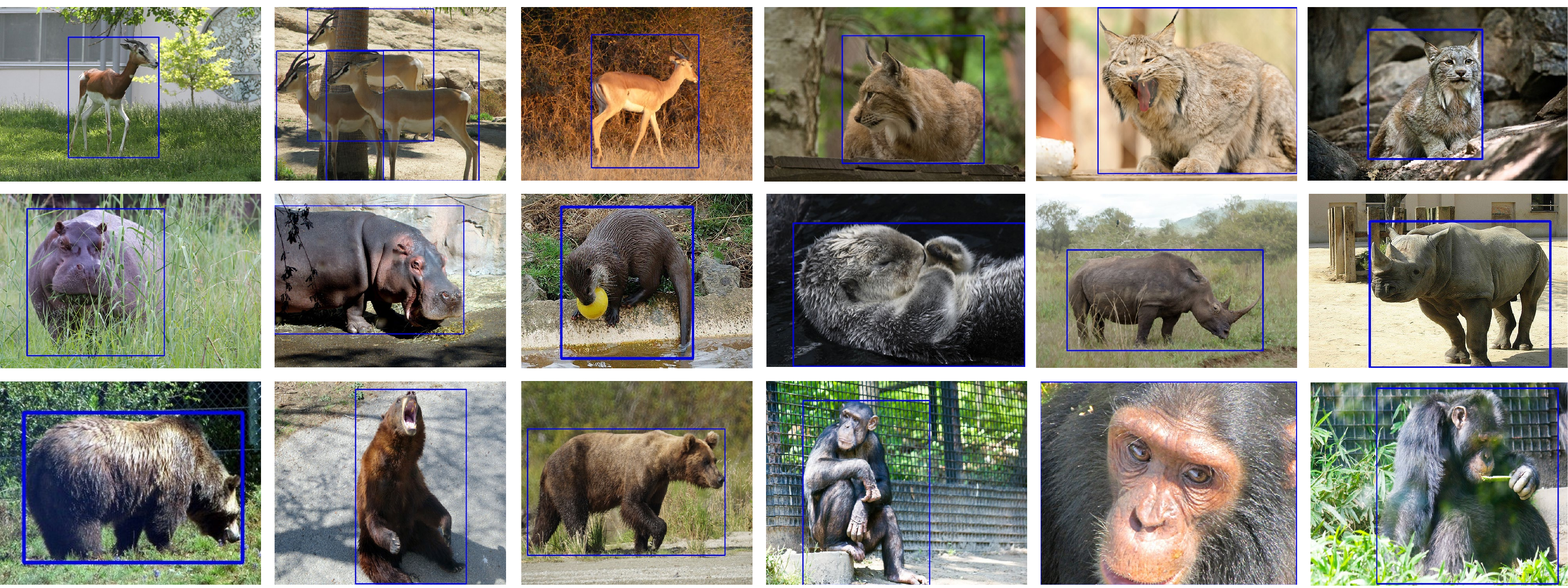}
        \caption{Samples of seven novel animal categories with bounding box provided in our dataset.}
        \label{7animalbbox}
    \end{figure*}

    To bring convenience for the attempt to do domain adaptation of animal pose estimation to novel animal categories, we also provide bounding-box-labeled data of seven novel animal categories: otter, antelope, bear, chimpanzee, rhino, bobcat, and hippopotamus, in our proposed dataset. We select some samples in Fig \ref{7animalbbox}.

\begin{figure}[!htp]
        \centering
        \includegraphics[width=\linewidth]{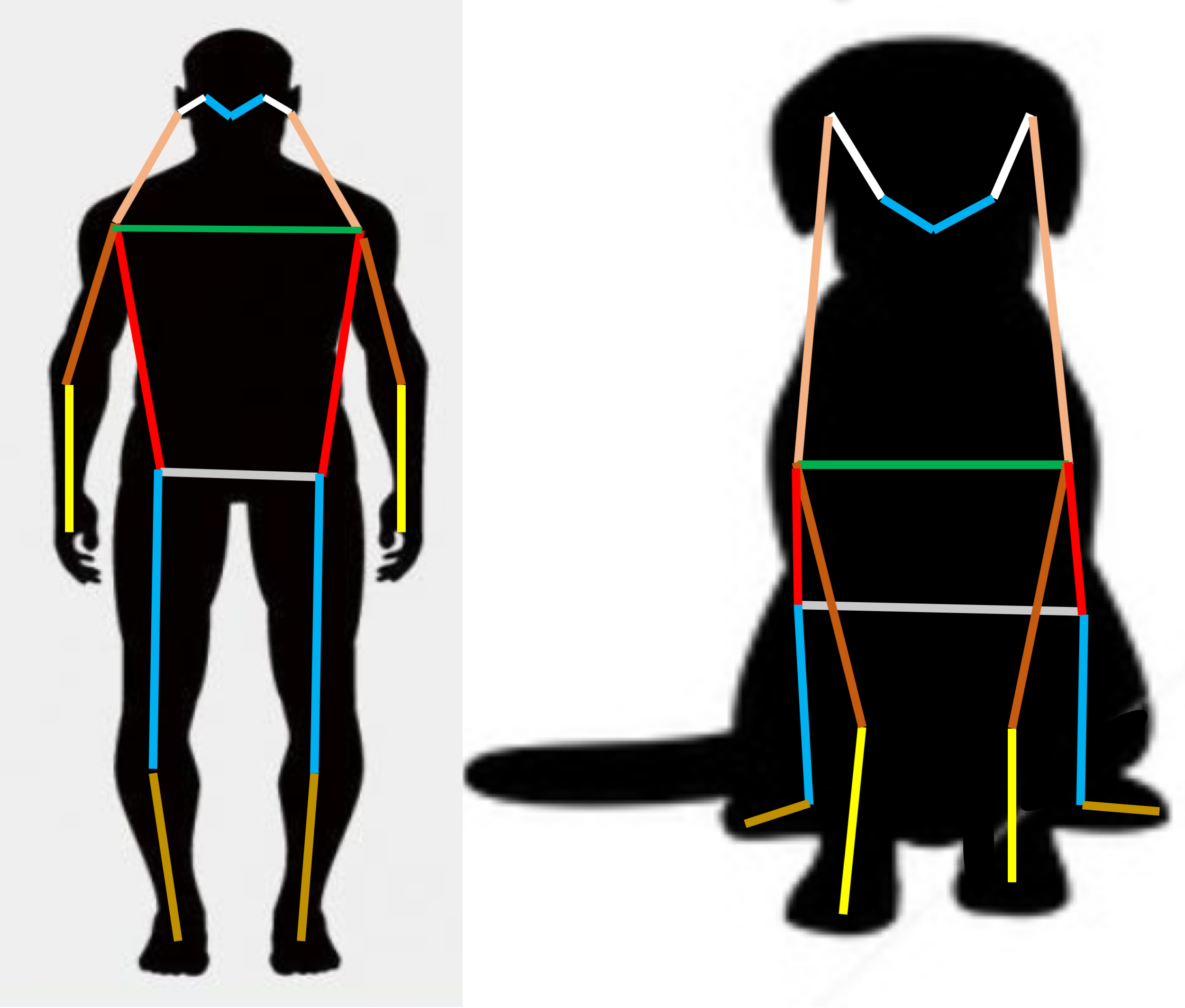}
        \caption{18 'bones' defined based on COCO-format keypoint annotations.}
        \label{bones}
    \end{figure}

    As explained in our paper, we use 18 pre-defined 'bones' to evaluate the domain shift on keypoints between animals and humans, the definition of which is shown in Fig \ref{bones}. 
    
    \section{Failure Cases of Proposed Methods}
    
\begin{figure*}[!htp]
       \includegraphics[width=17.5cm]{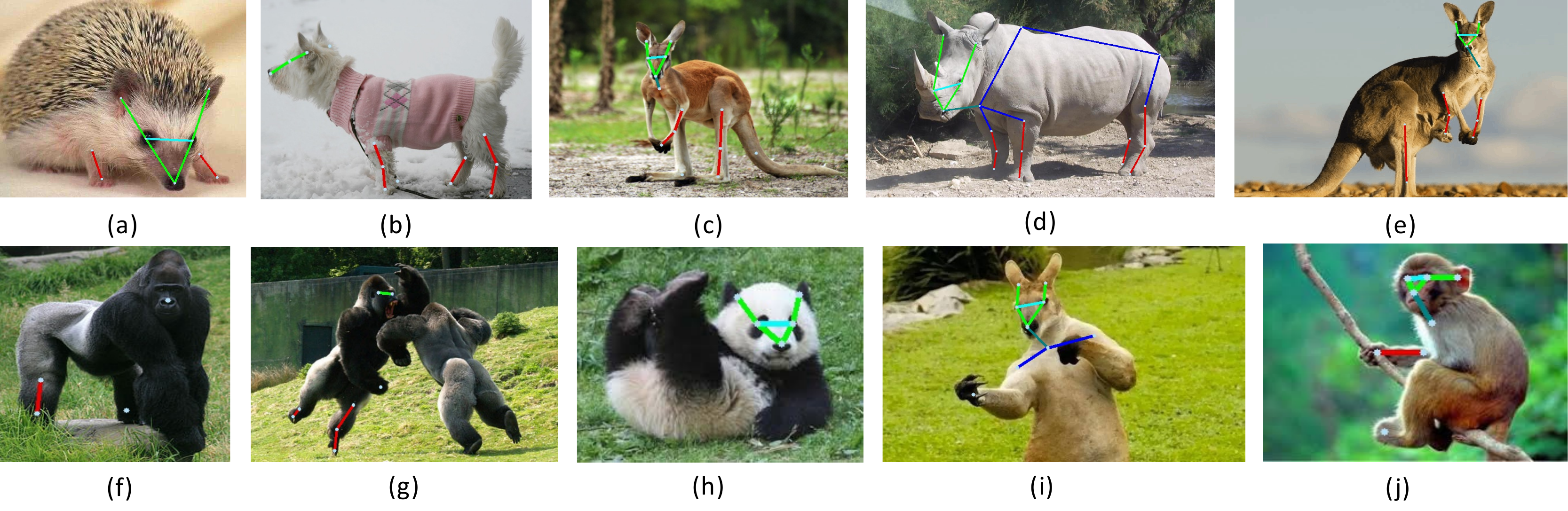}
       \caption{Samples of failure cases generated by our proposed methods on unseen animal categories.}
    \label{failure1}
    \end{figure*}

    We also sampled some representative failure cases in Fig \ref{failure1}. Unusual appearance may make keypoints unrecognized. For instance, our model report unrecognized withers on hedgehog with spine on back(a) and dog with clothes(b). Rhino with horn(d) makes the estimation of face-keypoints total chaos. On the other hand, bad global feature such as too low contrast ((f),(g)) or unusual gesture ((h),(i),(j)) bring difficulty to our model as well.
    
    \section{How ground truth helps model performance}
    
    To compare with the reported model performance of unsupervised domain adaptation (Table 2 in the paper), we provide more annotations on the target domain into training trying to reach an accuracy upper bound. We annotate 200 instances of each category as extra supervision (which are also contained in the released dataset). The result is shown in Table \ref{upperbound}. It proves that even our proposed methods help a lot to do unsupervised animal pose estimation by domain adaptation, the domain shift between different animal categories still harms model performance very much. More intuitively, to introduce some ground truth on the target domain can greatly boost the model performance. To summarize, to achieve more reliable domain adaptation results and to make labor intensive labeling work less necessary, there is still much work to do. Another interesting fact found from the experiment is that the extent of the benefit model gains from introduced ground truth (performance gap between supervised and unsupervised settings) varies very much on different categories. We think it might result from the different domain shift between different domains.
    
    \section{Failure of GANs for Data Augmentation}

    Although some previous works\cite{cyclegandetection} report success on domain adaptation benefiting from data augmentation by GANs\cite{CycleGAN2017}, we found it hardly work on our task. There are two major reasons for that. 1) current GANs usually work well on style transfer tasks but get stuck in failure when the wanted image after the transfer is too different from the original images. For example, the popular CycleGan\cite{CycleGAN2017} can give promising performance when transferring white dogs to black dogs but show severe unstability when transferring a dog to cat, which is also reported in \cite{CycleGAN2017}. 2) as we only have pose label on animals of limited classes, the data augmentation is expected to transfer animal sample of pose-labeled class to unseen class and to keep the original labels valid. However, even GANs output some relatively successful results which are expected to do data augmentation for unseen animals. The keypoint labels on original instance are likely not to be accurate any more.
    
    \begin{table}
        \begin{center}
        \tabcolsep=2.0pt
        \begin{tabular}{@{}lccccc}
        \toprule
        & \multicolumn{5}{c}{mAP for each class}\\
        \cline{2-6}
        $N_{GT}$ & cat & dog & sheep & cow & horse\\
        \hline
        0 & 42.3 & 41.0 & 54.7 & 57.3 & 53.1\\
        50 & 71.2 & 67.0 & 64.0 & 60.1 & 68.5\\
        100 & 71.5 & 67.8 & 64.8 & 61.3 & 69.0\\
        200 & 72.7 & 68.4 & 66.7 & 64.6 & 70.3\\
        \bottomrule
        \end{tabular}
        \end{center}
        \caption{Comparisons of model performance in unsupervised manner and supervised manner. $N_{GT}$ is the number of instance with labeled ground truth label put into training. The training settings of all groups are as same as described in the paper.}
        \label{upperbound}
    \end{table}
    
    \begin{figure*}
       \includegraphics[width=18cm]{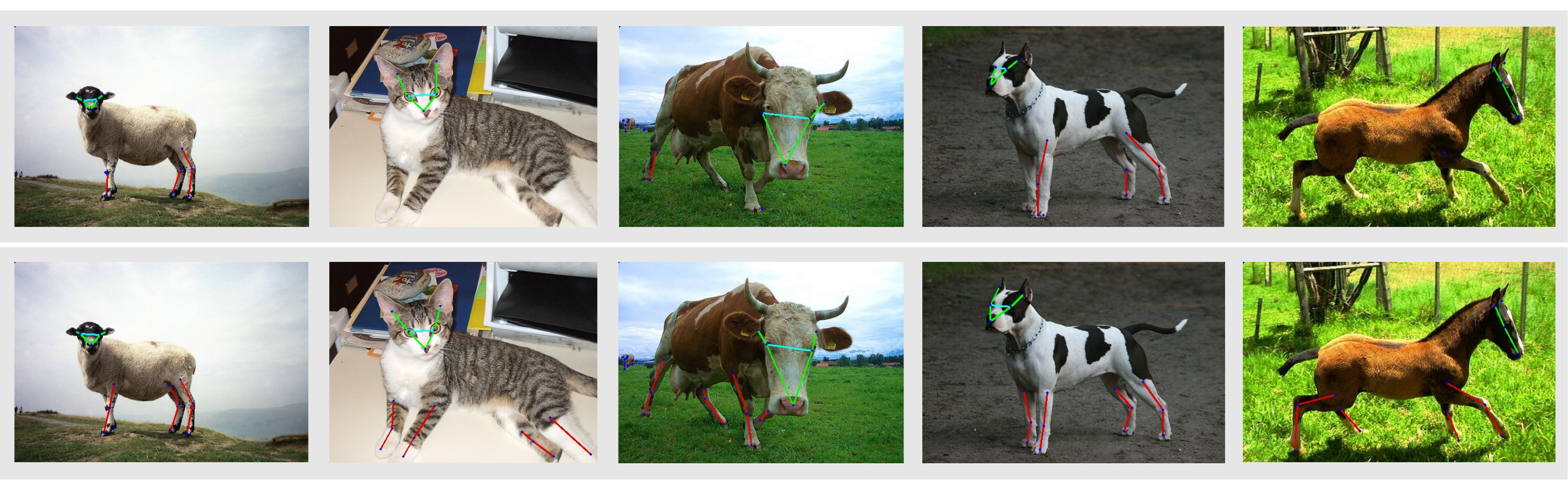}
       \caption{Upper images are estimated pose by model trained without WS-CDA. Lower ones are obtained after model being trained with WS-CDA. Common 17 keypoints between animals and humans are selected for visualization.}
        \label{WSCDA_improvement}
    \end{figure*}
    
    \end{document}